\documentclass[letterpaper, 10pt, conference]{ieeeconf}  

\IEEEoverridecommandlockouts                              

\usepackage[ruled,linesnumbered]{algorithm2e}
\usepackage{setspace, algpseudocode }
\usepackage{graphicx}
\usepackage{amsmath}
\usepackage{amssymb}
\usepackage{booktabs}
\usepackage[pagebackref,breaklinks,colorlinks]{hyperref}
\usepackage{url}
\usepackage{xspace}
\let\labelindent\relax
\usepackage{enumitem}
\usepackage[table]{xcolor}
\usepackage{booktabs}
\usepackage{multicol}
\usepackage{multirow}
\usepackage{graphicx}
\usepackage{array}

\usepackage{pifont}

\usepackage[capitalize]{cleveref}
\crefname{section}{Sec.}{Secs.}
\Crefname{section}{Section}{Sections}
\Crefname{table}{Table}{Tables}
\crefname{table}{Tab.}{Tabs.}

\newcommand{\ours}{SfA\xspace}
\newcommand{\oursfull}{Structure from Action\xspace}

\newcommand{\mypara}{\vspace{1mm}\noindent\textbf}
\newcommand{\etal}{\textit{et al.}}

\newcommand{\customfootnotetext}[2]{{
  \renewcommand{\thefootnote}{#1}
  \footnotetext[0]{#2}}}

\definecolor{grey}{rgb}{0.9,0.9,0.9}

\title{\LARGE \bf
Structure from Action:\\Learning Interactions for 3D Articulated Object Structure Discovery
}

\author{Neil Nie$^\diamond$\quad Samir Yitzhak Gadre$^\diamond$\quad Kiana Ehsani$^{\dag}$\quad Shuran Song$^\diamond$\\
\href{https://sfa.cs.columbia.edu/}
{https://sfa.cs.columbia.edu/}
}

\begin{document}

\maketitle

\customfootnotetext{$\diamond$}{
Columbia University, $^{\dag}$Allen Institute for AI.
Correspondence to \texttt{neil.nie@columbia.edu}.}

\thispagestyle{empty}
\pagestyle{empty}

\vspace{-15mm}
\begin{abstract}
We introduce Structure from Action (SfA), a framework to discover 3D part geometry and joint parameters of unseen articulated objects via a sequence of inferred interactions. 
Our key insight is that 3D interaction and perception should be considered in conjunction to construct 3D articulated CAD models, especially for categories not seen during training.
By selecting informative interactions, SfA discovers parts and reveals occluded surfaces, like the inside of a closed drawer.
By aggregating visual observations in 3D, SfA accurately segments multiple parts, reconstructs part geometry, and infers all joint parameters in a canonical coordinate frame.
Our experiments demonstrate that a SfA model trained in simulation can generalize to many unseen object categories with diverse structures and to real-world objects.
Empirically, SfA outperforms a pipeline of state-of-the-art components by 25.4 3D IoU percentage points on unseen categories, while matching already performant joint estimation baselines.\footnote{For code, data, and videos, see \href{https://sfa.cs.columbia.edu/}{\texttt{sfa.cs.columbia.edu/}}}

\end{abstract}

\section{Introduction}
\label{sec:introduction}

For robots to be useful out-of-the-box, they must handle a variety of objects---even those that are unfamiliar.
Beyond rigid objects, articulated objects, like drawers and microwaves, are of particular interest~\cite{abbatematteo_2020, mo_2021, Gadre2021ActTP}, especially in household use-cases.
For tasks involving novel articulated objects, recovering 3D articulated CAD models (e.g., URDFs) is a promising starting point, as they are immediately useful in task-specific planning pipelines~\cite{Sturm2010OperatingAO,Burget2013WholebodyMP,Capitanelli2017AutomatedPT,Capitanelli2018OnTM,Mittal2021ArticulatedOI}. 
For instance, recovering models of kitchen drawers can lay the foundation for downstream planning to retrieve objects within them.
To discover the structure of objects beyond training categories, there is evidence that interaction is critical~\cite{Gadre2021ActTP,xu2022umpnet}.
Informative interaction allows an agent to expose kinematic constraints (e.g., prismatic or revolute joints) and observe occluded part geometry.

Inferring joints, kinematic constraints, and the full 3D structure of articulated objects is a complex task that involves tackling a diverse set of challenges:
\begin{itemize}[leftmargin=3mm]
    \item \textbf{Inferring informative interactions.} Given unstructured point clouds, an agent must act intentionally to expose structures, as
    random actions and repetitive actions may not give signal about articulation.
    \item \textbf{Persistent part aggregation in 3D.} From an observed sequence of interactions, it is necessary to discover new parts and track existing parts, even in the presence of severe occlusion.
    If an agent closes a drawer, the part should persist within the object representation, even when it is not directly visible in the following steps. 
    \item \textbf{Cross-category generalization.} The algorithm should handle object categories unseen during training, with different joint configurations.
\end{itemize}

\begin{figure}[t]
\includegraphics[width=0.98\linewidth]{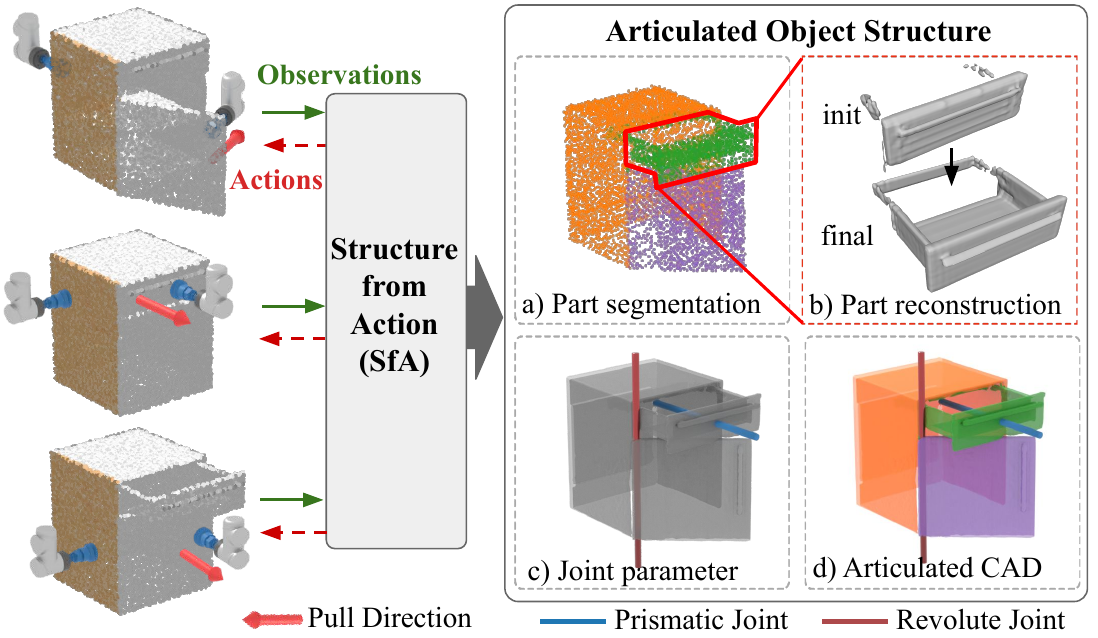}     \vspace{-2mm}
    \caption {\textbf{\oursfull.}
    Our framework discovers an object's structure through a sequence of 3D interactions.
    The resulting structure includes a) part segmentation, b) 3D reconstruction for each part, and c) joint parameters, together describing d) a 3D articulated CAD model.}
    \label{fig:overall}
    \vspace{-3mm}
\end{figure}

These challenges have motivated simplifying assumptions in prior works (e.g., objects lie flat \cite{Gadre2021ActTP} or interactions are given \cite{jiang2022ditto}).
In this work, we introduce an approach for constructing articulated 3D CAD models of objects using interactions, thereby relaxing the aforementioned assumptions.

To address these challenges, we introduce \emph{\oursfull (\ours)} to expose the object parts and joints through interaction. 
Our \emph{key insight} is that 3D interaction and perception must be considered in conjunction to construct 3D articulated CAD models.
Specifically, \ours learns 1) a sequential interaction policy to expose the object's hidden part geometry and kinematics, 2) a dynamic part reconstruction module that segments and completes the object parts by aggregating visual observations, and 3) a joint estimation module that infers object joint types and parameters based on the observed motion. 
The final output is a 3D articulated CAD model (see Fig.~\ref{fig:overall}).

We evaluate \ours on unseen object instances and categories from the PartNet-Mobility~\cite{chang_2015, Mo_2019, Xiang_2020} dataset.
Our experiments validate the following contributions:
\begin{itemize}[leftmargin=3mm]
    \item An interaction policy that learns informative interaction strategies in 3D to recover 3D articulated object structure.
    
    \item A learnable perception module that aggregates visual observations on-the-fly to improve the accuracy of part reconstruction and joint estimation. 
    
    \item A single \ours model (both the interaction and perception modules) trained in simulation can generalize to many unseen object categories with unknown kinematic structures, and to real-world objects. 
\end{itemize}

\section{Related Work}
\label{sec:related}

Recently, interactive perception with articulated objects has gained renewed interest.
Here the goals are to recover objects' articulation structure, including objects' part reconstruction, segmentation, and joints estimation. An algorithm should also handle objects with multiple parts.
While prior work tackles some of these challenges, \ours presents a comprehensive framework addressing all facets of the problem.

\mypara{Articulated object manipulation.} Articulated objects are an important class of objects for manipulation, and the community has come a long way to make datasets and benchmarks to facilitate research in this direction~\cite{chang_2015, Mo_2019,MartnMartn2019TheRD,Xiang_2020,maniskill2021,Liu2022AKB48AR}.
A line of work tackles the problem of interacting with articulated objects to move their parts~\cite{mo_2021,Wu2021VATMartLV,Mittal2021ArticulatedOI,Shen2022LearningCG}. Some work \cite{Liu2021VMAOGM,Bertolucci2021ManipulationOA} uses dual-arm manipulators to enable more complex interaction. This work mostly focuses on interacting with the purpose of completing a high-level task (such as opening cabinets~\cite{Shen2022LearningCG}, etc.).
Our goal is to learn to interact with objects to discover joints and parts.
Eisner, \etal~\cite{Eisner2022FlowBot3DL3} propose a vision-based method to predict the flow and articulated motions of an object. However, they do not infer part segmentation or joints.
Xu, \etal~\cite{xu2022umpnet} propose a single image-based policy network to recover joint axes, but do not attempt to recover parts. 

\mypara{Perception from passive observation.}
Prior work has used a variety of methods to recover object joint constraints, such as using dense pose fitting~\cite{Desingh2019EfficientNB}, adapting neural radiance field~\cite{Noguchi_2021_ICCV}, inferring kinematic graphs~\cite{AbdulRashid2021LearningTI}, and semantic segmentation~\cite{Weng2021CAPTRACP}. Mu, \etal~\cite{Mu2021ASDFLD} propose a model to generate shapes of articulated objects at unseen angles. These methods require prior knowledge of the object or are category-dependent.
Moreover, researchers have addressed the part segmentation and structure recovery from non-sequential data (e.g., a single view or point cloud)~\cite{zeng2021visual, Wang_2015, TsogkasKPV15, hung2019scops, Lee2020CameratoRobotPE, abbatematteo_2020, li2020categorylevel, qi2017pointnet, qi2017pointnet2, Yue_2019, huang_2021}.
In contrast, our method uses a sequence of data, which enables discovering parts of unseen object categories without prior knowledge.
The community has tried to recover and track object structures from motion cues between sequential observations~\cite{Jain2021DistributionalDE, jain2021screwnet, liu2020nothing, black1998eigentracking,Yan_2006, xu2019Unsupervised, Yi_2019, Schmidt_2014, martin_2014,Siarohin_2021_CVPR,noguchi2021watch,yang2021viser,Qian2022Understanding3O,jiang2022ditto,zeng2021visual}. However, these methods rely on motion existing in the scene. Our method uses previous observations to predict actions that result in informative motions. 

\mypara{Perception from interaction.}
Classical approaches use hand-tuned actions to create informative motion for downstream perception~\cite{Sturm_2011,Katz_2013,Pillai_2014}. In contrast, we use a generalizable approach to predict the actions, even for novel categories.
Similarly, more modern approaches focus on perception, using scripted robot actions and considers only a single interaction timestep~\cite{jiang2022ditto}. 
Kumar, \etal~\cite{Kumar2019EstimatingMD} recover the mass distribution of the articulated objects using interaction, but they do not recover joints or parts.
Gadre, \etal~\cite{Gadre2021ActTP} proposed a method that learns both interaction and perception.
However, they consider a simplified 2D case with revolute joints.
In contrast, by using 3D actions and perception, we are able to consider both revolute and prismatic joints and relax restrictions on camera positioning.
Lv, \etal~\cite{lv2022sagci} proposed SAGCI, an interactive perception method for articulated object structure discovery using a differentiable physics engine. However, it does not explicitly complete part geometry, nor does it represent occluded part geometry persistently, which are core functionalities supported by \ours.
More recently, Hsu \etal~\cite{Hsu2023DittoITH} propose Ditto in the House, which extends Ditto~\cite{jiang2022ditto} for discovering many parts and joints in scenes, leveraging a learned interaction policy.
However, unlike \ours, they do not condition their policy on their perceptual inference, thereby breaking the perception-interaction loop.
While they show a human-interaction proof of concept, we implement a real-world \ours with a UR-5 robot and RealSense cameras.
\begin{figure*}[t]
\includegraphics[width=0.98\linewidth]{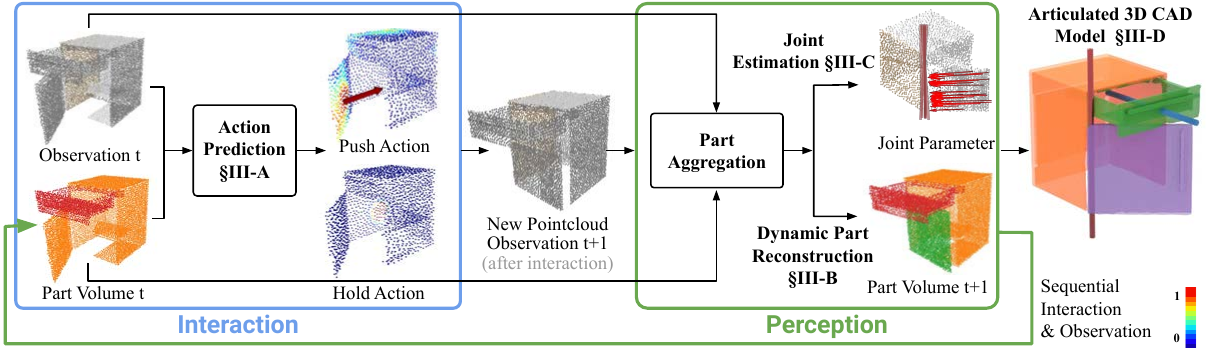} \vspace{-2mm}
    \caption {\textbf{Overview.} Given an RGB point cloud observation of an unknown articulated object, \ours infers and executes a sequence of informative actions (\S~\ref{sec:policy}), discovers and reconstructs parts (\S~\ref{sec:part}), estimates joint parameters (\S~\ref{sec:joints}), and outputs an articulated 3D CAD model of the object (\S~\ref{sec:urdf}).}
    \label{fig:approach}
    \vspace{-3mm}
\end{figure*}

\section{Structure From Action}
\label{sec:approach}

We introduce \oursfull (\ours), a learning framework to interact with articulated objects to discover their parts and joints.
Our framework is agnostic to object category and to the number of parts and joints that constitute an object.
Hence, \ours can generalize to novel categories.
Given an observation $P_0$, the initial RGB point cloud before any interaction, \ours infers actions to reveal an objects' parts and joint structure (\S~\ref{sec:policy}). Then by observing the object motion, \ours discovers and reconstructs the object part using a part aggregation module (\S~\ref{sec:part}) and infers joint parameters using a joint estimation module (\S~\ref{sec:joints}). Over several timesteps, the output of the algorithm is an articulated CAD model consisting of 3D part meshes along with the revolute and prismatic joints that connect them (\S~\ref{sec:urdf}). 
Fig. \ref{fig:approach} gives an overview of our approach.

\subsection{Learning to Interact with Articulated Parts}
\label{sec:policy}

The first step of \ours is to infer informative actions to reveal an object's kinematic structure. An action is informative if it isolates an individual part, instead of moving the whole object or multiple parts.
Furthermore, an informative action should attempt to move new parts, instead of interacting with the same part repeatedly.

\mypara{Action representation.} 
Inspired by AtP \cite{Gadre2021ActTP}, we assume a robot with two arms, which uses its end-effectors to simultaneously hold and push different parts of the object. These interactions allow the agent to isolate a single part of the object and are particularly useful for small objects without a fixed base.
However, unlike AtP, we consider a \emph{continuous 3D action space} instead of a discrete 2D action space.
This provides the flexibility to handle parts rotating and sliding about arbitrary axes. 
We represent a hold action as a 3D point location. 
We define a push to be a 3D point location and 3D direction along which an agent applies a fixed force. Note, this definition makes no distinction between ``pulling" and ``pushing". In terms of the mechanics of pulling, we assume that the agent has access to a suction gripper that can be used to, say, pull a drawer or push a door. 

\begin{figure}[t]
\centering
\includegraphics[width=0.48\textwidth]{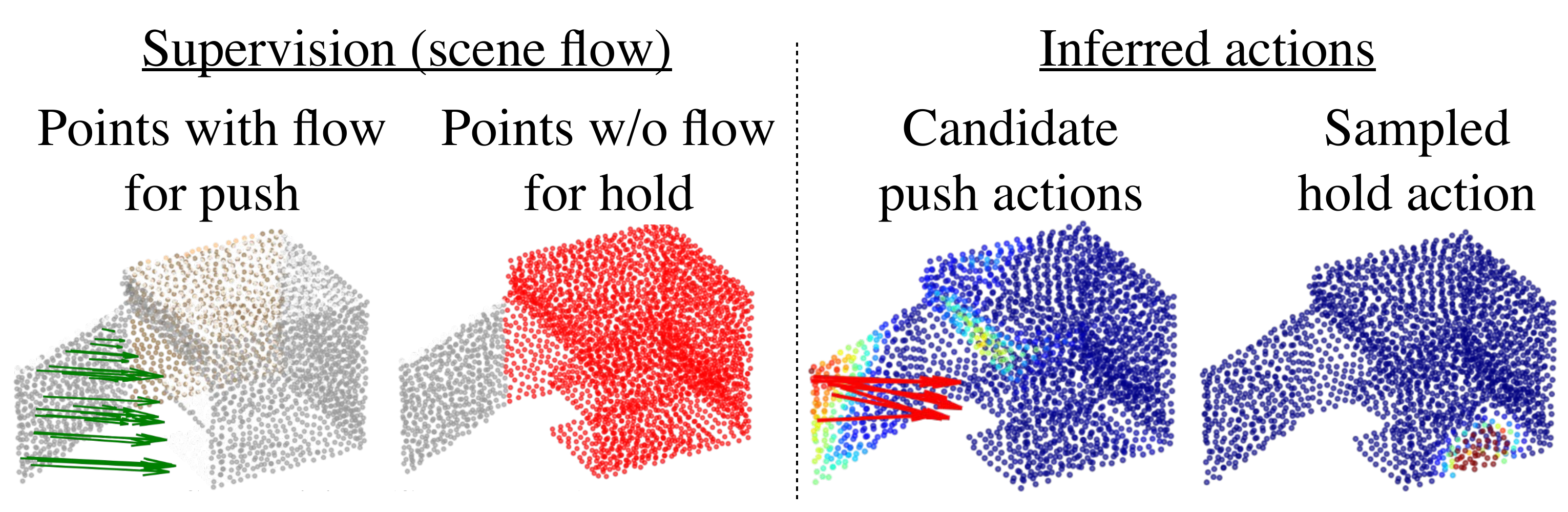}
    \caption {\footnotesize \textbf{Learning Interaction Policy.} 
    \emph{(left)} During training, the 3D scene flow is used to supervise the action directions (green). For a timestep, areas where flow is zero are assumed to be good hold locations (red).
    \emph{(right)} Inferred candidate push actions conditioned on a sampled hold action.
    }
    \label{fig:policy}
    \vspace{-5mm}
\end{figure}

\mypara{Action inference.}
The input to the action inference module is the current object observation point cloud $P_t$ and part history voxel volume $\mathcal{H}_t$. $P_t \in \mathbb{R}^{n \times c}$ is formed by selecting $n$ points via farthest point sampling over posed RGBD images.
In our case $n=2048$.
We consider $c = 9$ channels encoding the point's XYZ location, 3D surface normal, and RGB color. 
A part history volume $\mathcal{H}_t$ encodes the agent's current belief about the object's part segmentation and is spatially aligned with $P_t$ (see \S\ref{sec:part} for more details on $\mathcal{H}_t$).
We wish to associate each point with its current segmentation prediction.
Hence, we concatenate each point in $P_t$ with its corresponding value from $\mathcal{H}_t$, before passing the points into the action inference module. 
Action inference is hence conditioned on the current belief about the part segmentation.
Intuitively, we want inferred actions to push parts that are not already confidently segmented so that the downstream perceptual model (\S\ref{sec:part}) is able to discover these new parts.

The action inference module is composed of two point transformer encoder-decoders~\cite{Zhao2020}, the first to infer a hold score for each point and the second to infer a push action for each point conditioned on a sampled hold location.

To predict the hold action, the network infers a score for every point.
A higher score indicates a better hold location.
We sample a hold location uniformly over the top $k=100$ hold scores.
We do not want to push on a part that we are already holding.
Hence, we condition the push prediction on the selected hold action. Concretely, for each point in $P_t$, we compute the point-wise distance to the selected hold location and use it as an additional input to the push network. 
The push network outputs a flow vector for each input point, where the vector directions (seen in Fig. \ref{fig:policy} \emph{(right)}) indicate the inferred push directions and the magnitude indicates the push score of the action. At inference, we select the push with the highest score to execute in tandem with the hold.

\mypara{Dataset creation.}
3D scene flow on a part can imply effective push actions on that part~\cite{Eisner2022FlowBot3DL3}.
The direction of a good push action is aligned with flow vectors, while the magnitude of each flow vector gives a notion of how effective a push is.
Take for instance a door that swings open. Locations with larger flow vectors correspond to points farther away from the revolute axis.
Interacting with such points is more likely to create discernible motion given a push action with a fixed force.
We also notice that points with no flow can be used as candidates for the hold action.
While all points without flow are not always equally good for holding, our results suggest that this approximate supervision is sufficient in practice.

Based on this intuition, we generate a supervised dataset using the PyBullet~\cite{coumans2016pybullet} simulator and URDF assets from PartNet-Mobility~\cite{Mo_2019}. 
We move a single part per step by changing its simulation joint state directly.
Once a part has moved we consider it \emph{discovered}.
We repeat this process for five timesteps per object, moving parts that have not been discovered before moving parts that have already moved.
At each timestep, we save the point cloud generated from posed RGBD views, observable scene flow per point, and the ground truth part labels, with a single label for undiscovered parts and unique labels for each discovered part. 
Once a part has moved, we generate a categorical label for it.

\mypara{Supervision and training.} 
Recall our model takes the current point cloud observation and the current part history segmentation as input, it then predicts hold scores per point, samples a hold location and predicts push scores per point conditioned on the hold location.
During training, we sample interactions from our dataset i.i.d.
By using 3D flow as supervision as in Fig.~\ref{fig:policy} \emph{(left)} and the ground truth history as input, we supervise the hold network to predict no-flow points with binary cross-entropy loss.
The push network is trained to predict 3D scene flow using MSE loss. 
Note, ground truth history is used for \emph{training only}.

\begin{figure}
\includegraphics[width=0.44\textwidth]{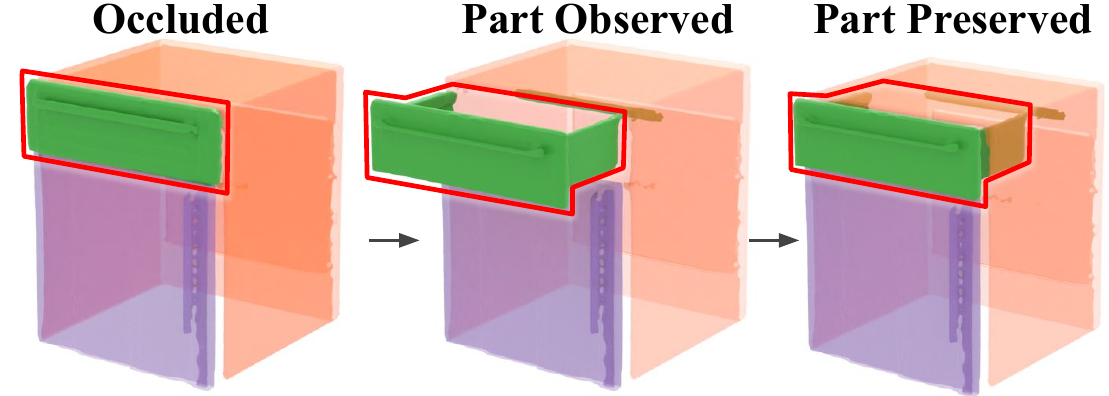} 
\caption{\footnotesize \textbf{Dynamic part reconstruction.} \ours completes part geometry by aggregating all past observations in a spatially consistent manner. }
\vspace{-5mm}
\label{fig:dynamic_part}
\end{figure}

\subsection{Learning Persistent Part Aggregation}
\label{sec:part}

The goal of the part aggregation module is to construct a history volume $\mathcal{H}_t$ that encodes the agent's current belief of the object structure (i.e., segmentation and geometry) from all the past observations. Performing such part aggregation is challenging since it requires the algorithm to establish reliable correspondences between the part before and after the movement. Here, point-to-point correspondences are insufficient as large portions of the surface may disappear (e.g., a drawer as it closes).
To tackle these challenges, we propose a learning-based part aggregation module.

We choose to use volumetric representation to allow the network better leverage the spatial alignment between different observations and the history volume $\mathcal{H}_t$. 
We represent $\mathcal{H}_t \in \mathbb{R}^{v \times v \times v \times d}$, which is a 3D segmentation volume aligned to the current observation in the world frame. 
In our case, $v=96$, representing spatial dimensions and $d=7$ is the channel dimension.
The $d$ channels store a probability distribution over part indices, with the first channel representing free space.
Intuitively, $\mathcal{H}_t$, can be decoded to a discrete segmentation by taking $\max_d$ at each voxel.

$\mathcal{H}_0$ is initialized with all occupied voxels from the initial point cloud observation assigned to the first part with probability one.
Over a few interactions, we want to update $\mathcal{H}$ to more accurately capture the various parts that make up the object.
If a discovered part (say $i$-th part) gets moved again, the part aggregation module should update the occupancy of $i$-th channel in $\mathcal{H}$ to reflect new observations, like filling in surfaces that were previously occluded (Fig.~\ref{fig:dynamic_part} 2nd step) or preserving geometry when it is moved into occlusion (Fig.~\ref{fig:dynamic_part} 3rd step).
The model must also learn to copy over labels of stationary parts to maintain parts' permanence across interaction steps. 

\mypara{Part aggregation network.}
The aggregation network is constructed as 3D CNN.
It takes the history $\mathcal{H}_{t-1}$ and voxelized point clouds $V_{t-1},V_{t} \in \mathbb{R}^{v \times v \times v \times 7}$ as input, and outputs a new history $\mathcal{H}_{t}$.
The $7$ channels encode the object's occupancy (1D), surface normal (3D) and color (3D).

\mypara{Supervision and training.} 
We construct the target history volume $\mathcal{H}_{t}^\textrm{gt}$ together with the offline data generation process described in \S \ref{sec:policy}.  
At each step $t$, the target volume includes channels for the parts moved by the agent and allocates new channels if new parts are observed. For each part channel, the target volume will include all surfaces that the camera has observed in any of the past and current steps $\in (0,t]$, including surfaces that get occluded in this step.

Since $\mathcal{H}_{t}^\textrm{gt}$ is generated with a consistent part index across steps, the network learns to keep track of part identity over multiple interaction steps after the part was first discovered, without explicitly tracking parts.
Moreover, since $\mathcal{H}_{t}^\textrm{gt}$ introduces part geometry incrementally for each step (only after the surface is observed).
It allows the network to learn how to ``aggregate'' existing observations without the need to ``guess'' the unobserved part geometry.
Finally, since the $\mathcal{H}_{t}^\textrm{gt}$ preserves the part geometry once it is observed, it allows the network to learn object permanence during occlusion. As a result, this part aggregation module is able to discover, track, and reconstruct the object part geometry using a single network.
The network and trained with voxel-wise cross-entropy loss between the predicted and target volume.

\subsection{Recovering Joints}
\label{sec:joints}

\begin{figure}[t]
\includegraphics[width=0.44\textwidth]{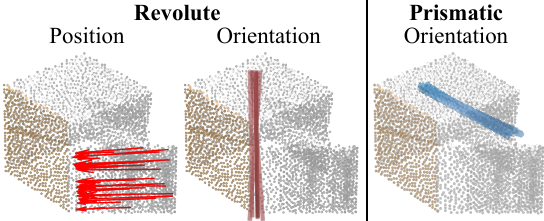}
\caption{\footnotesize\textbf{Joint Inference.} \emph{(left)} Revolute joint position and axis orientation votes. \emph{(right)} Prismatic joint orientation votes.}
\vspace{-5mm}
\label{fig:joint_inference}
\end{figure}

Apart from the part information, it is also critical to infer the object's joint parameters to fully recover its kinematic structure. To do so, we designed a joint inference module that infers the object's joint type and parameters from two consecutive object observations $P_{t-1}$ and $P_{t}$ with object motion. If no part has moved, this interaction step will be ignored for joint prediction.

With the learned action policy (i.e., simultaneously holding and pushing different parts), the agent tries to move a single part at each step.
This interaction strategy greatly simplifies the joint inference module, which only needs to consider the case of one component moving about a joint.

If more than one part is moved, the model will treat all moving parts as one common part and predict one set of joint parameters, this error could be fixed with future interaction steps. Lastly, we assume that all movable parts are connected to the base link via a joint, with the base link always labeled as part on in the segmentation volume.

\mypara{Joint network training and inference.}
The joint inference module (modeled as a 3D CNN) is inspired by prior work \cite{jiang2022ditto} and a popular joint parameter representation \cite{li2020categorylevel}.
This network takes as input $V_{t-1}, V_{t}$, which are the successive voxelized point clouds also considered by the part aggregation network.
The inferred joint parameters are represented as one volumetric output $J$ with three components: 1) $ J_\mathrm{type} \in \mathbb{R}^{v \times v \times v \times 1} $ for joint type trained with BCE loss. 
2) $ J_\mathrm{axis} \in \mathbb{R}^{v \times v \times v \times 3} $, gives per voxel predictions of the joint axis direction (seen in Fig. \ref{fig:joint_inference} \emph{(right)}), trained with cosine similarity loss with ground truth value.  
3) $ J_\mathrm{pos} \in \mathbb{R}^{v \times v \times v \times 1} $, gives per voxel predictions of the position of the revolute joint axis, which is represented using the distance between each voxel to its corresponding joint axis position (seen in Fig. \ref{fig:joint_inference} \emph{left}), trained with MSE loss. 

During training, we use the ground truth volumetric part labels and only supervise on the output voxels of the moved part. From these predictions, we can compute the joint parameters by averaging the predictions over all voxels labeled as the moving part inferred by the part aggregation module. 
To track multiple joints over several steps, we maintain a dictionary where the key is the part label inferred by the part aggregation model and the value is a list of $\{J_\mathrm{type}, \;\; J_\mathrm{axis}, \;\;  J_\mathrm{pos}\}$. If the policy interacts with a part more than once, the inferred joint parameters will be appended to the existing list in the dictionary.
The final joint parameters will be the median of all inferred values over several interaction steps.

\begin{figure*}[t]
    \centering
    \includegraphics[width=0.99\linewidth]{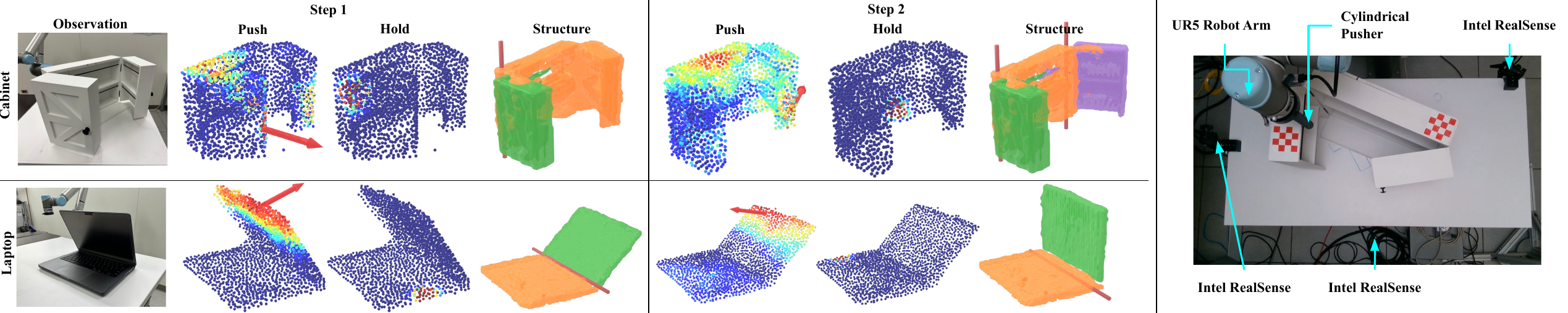} 
    \caption{\textbf{Realworld Result.} We evaluate the \ours pipeline on real-world point cloud constructed from multiple RGBD frames. The model performs well on previously unseen instances in the real world despite challenging noise artifacts from the real RGBD camera.}
    \label{fig:result_real}
\end{figure*}

\begin{figure*}[t]
    \centering
    \includegraphics[width=\linewidth]{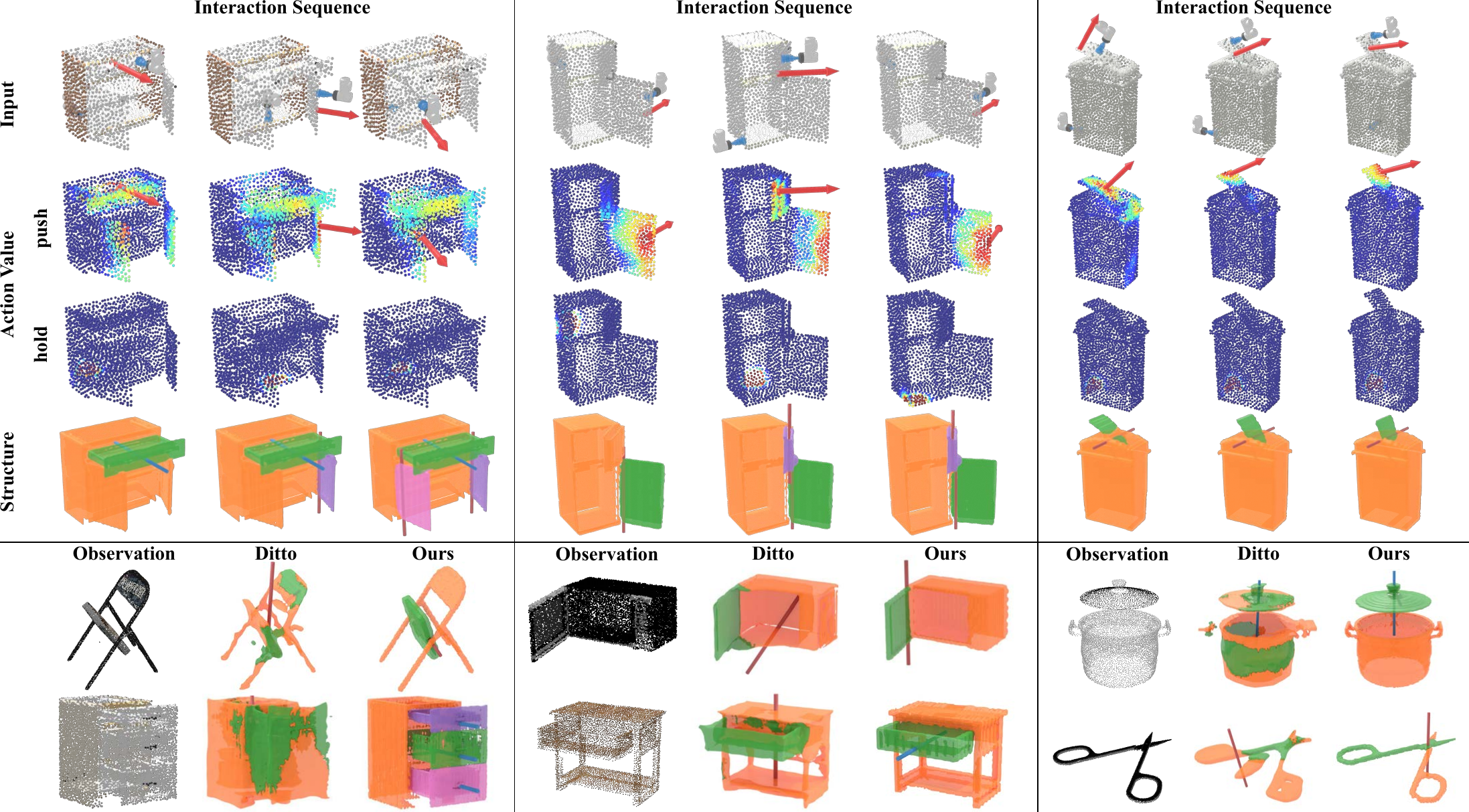}
    \caption{\textbf{Qualitative Result in Simulation.} We show the step-by-step results from the \ours pipeline. The inferred actions prioritize new parts discovery and expose articulations.  Our method outperforms the Ditto \cite{jiang2022ditto}  on both parts reconstruction and joints estimation (revolute: red, prismatic: blue).}
    \label{fig:result_sim}
    \vspace{-2mm}
\end{figure*}

\subsection{Constructing an Articulated CAD Model}
\label{sec:urdf}

Given the updated history volume $\mathcal{H}_t$, the last step is to extract the 3D mesh for each part.
Recall that each spatial entry in $\mathcal{H}_t$ encodes a probability distribution over parts.
We observe that computing an $\mathsf{argmax}$ over $\mathcal{H}_t$ can result in artifacts.
To circumvent this problem, we directly deal with the continuous probability values to extract a smoother surface. 
First, we compute the inverted probability volume $\hat{\mathcal{H}_t} = 1 - \mathcal{H}_t$, where a value closer to 0 indicates higher probabilities of the surface. Treating $\hat{\mathcal{H}_t}$ as a distance volume, we can apply marching cubes to extract surfaces. 
Since $\hat{\mathcal{H}_t}$ consists of continuous value, we can further upsample the volume (i.e., from $96^3$  to $288^3$) to improve the mesh quality without resorting to an expensive implicit surface representation. 
Finally, by combining the 3D part mesh with the estimated joint parameters (\S \ref{sec:joints}), we can generate a consolidated URDF file describing the articulated 3D CAD model as visualized in Fig.~\ref{fig:overall}(d)).

\begin{table*}[t]
{\footnotesize 
\centering
\setlength\tabcolsep{ 4 pt}

\caption{\label{tab:action} \textbf{Interaction policy evaluation.} 
} 
\vspace{-3mm}
\begin{tabular}{l|cccccccccc|cccccccc}
\toprule
     & \multicolumn{10}{c|}{Unseen Instances in Training Categories } &  \multicolumn{7}{c}{Unseen Categories}  \\ 
      & 
      \includegraphics[width = 0.035\linewidth]{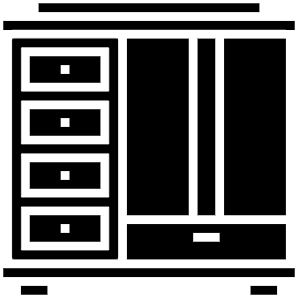} &
      \includegraphics[width = 0.035\linewidth]{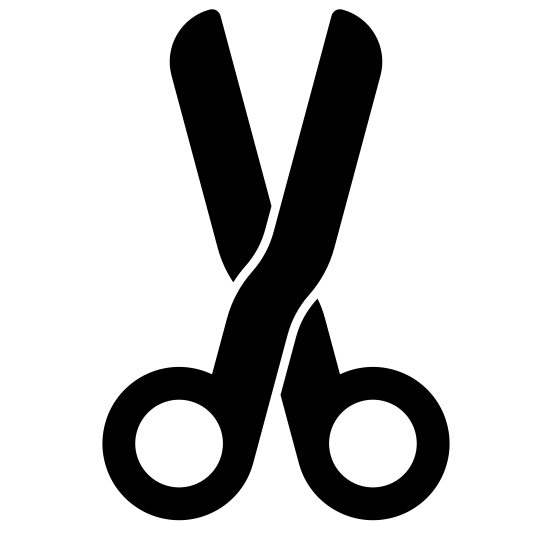} &
      \includegraphics[width = 0.035\linewidth]{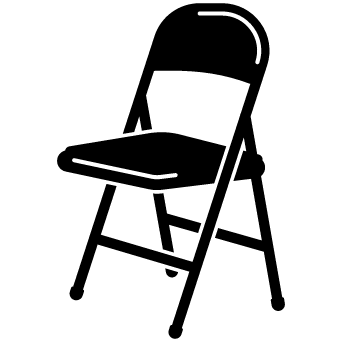} &
      \includegraphics[width = 0.035\linewidth]{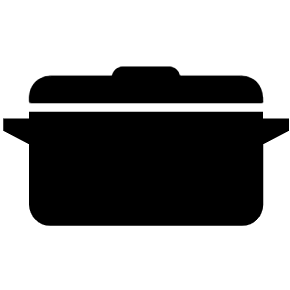} &
      \includegraphics[width = 0.035\linewidth]{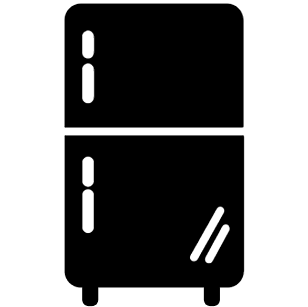} &
      \includegraphics[width = 0.035\linewidth]{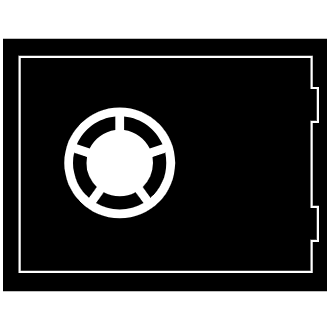} &
      \includegraphics[width = 0.035\linewidth]{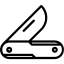} &
      \includegraphics[width = 0.035\linewidth]{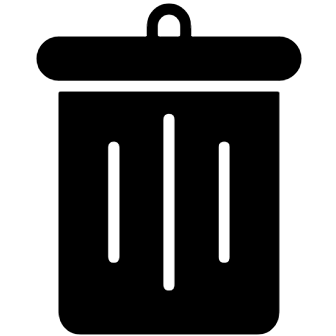} &
      \includegraphics[width = 0.035\linewidth]{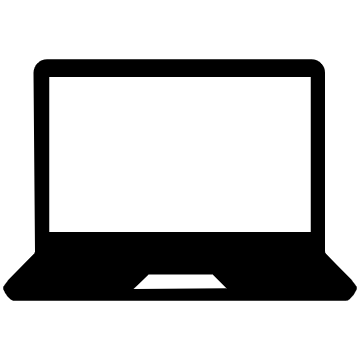} &
      \includegraphics[width = 0.035\linewidth]{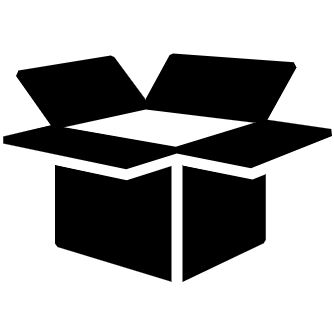} &
      \includegraphics[width = 0.035\linewidth]{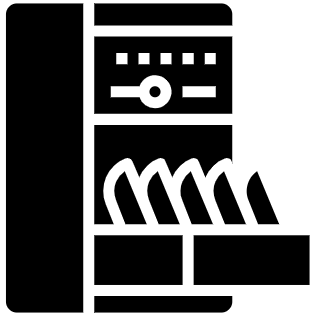} &
      \includegraphics[width = 0.035\linewidth]{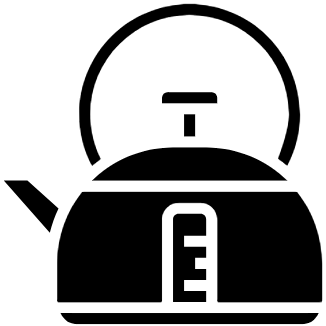}  &
      \includegraphics[width = 0.035\linewidth]{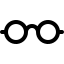} &
       \includegraphics[width = 0.035\linewidth]{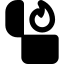} &
      \includegraphics[width = 0.035\linewidth]{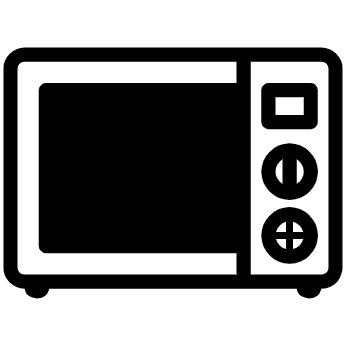} &
      \includegraphics[width = 0.035\linewidth]{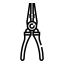} &
      \includegraphics[width = 0.035\linewidth]{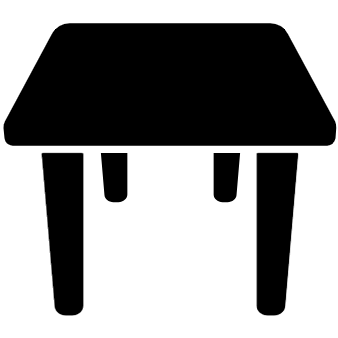} 
      \vspace{-0.2mm}
     \\

\midrule

AtP   \cite{Gadre2021ActTP} & 0.0 & 25.0 & 0.0 & 50.0 & 20.0 & 25.0 & 0.0 & 20.0 & 0.0 & 0.0  &  20.0 & 0.0 & 20.0 & 20.0 & 20.0 & 20.0 & 0.0  &  \\  
UMP-Net   \cite{xu2022umpnet}  & 0.0  & 50.0 & 10.0 & 0.0 & 18.2 & 28.5 & 0.0 & 0.0 & 70.0 & \textbf{83.3} &  6.2 & 16.6 & \textbf{22.7} & 26.3 & 5.5 & 60 & 5.3 \\  

\ours    & \textbf{60.0} & \textbf{61.6} & \textbf{77.5} & \textbf{100}& \textbf{56.6} & \textbf{95.0} & \textbf{90.0} & \textbf{66.6} & \textbf{86.0} & 73.3 & \textbf{83.7} & \textbf{51.6} & {19.1} & \textbf{65.4} & \textbf{86.6} & \textbf{70.0} & \textbf{22.2}  \\ 
\bottomrule
\end{tabular}

}
\end{table*}

\begin{table*}[t]
{\footnotesize 
\centering
\setlength\tabcolsep{ 3 pt}

\caption{\label{tab:part-discovery} \textbf{Part segmentation and reconstruction results}. } 
\vspace{-1mm}
\begin{tabular}{l|cccccccccc|cccccccc}
\toprule
     &\multicolumn{10}{c|}{Unseen Instances in Training Categories } &  \multicolumn{7}{c}{Unseen Categories}  \\ 
     &  
      \includegraphics[width = 0.035\linewidth]{icon/noun_Cabinet_2881254.png} &
      \includegraphics[width = 0.035\linewidth]{icon/noun-scissor-4857934.png} &
      \includegraphics[width = 0.035\linewidth]{icon/noun_folding_chair_2151184.png} &
      \includegraphics[width = 0.035\linewidth]{icon/noun_kitchen_pot_3363643.png} &
       \includegraphics[width = 0.035\linewidth]{icon/noun_Fridge_1875643.png} &
      \includegraphics[width = 0.035\linewidth]{icon/noun_safe_1202915.png} &
      \includegraphics[width = 0.035\linewidth]{icon/folding_knife.png} &
      \includegraphics[width = 0.035\linewidth]{icon/noun_trashcan_2244926.png} &
      \includegraphics[width = 0.035\linewidth]{icon/noun_Laptop_2291662.png} &
      \includegraphics[width = 0.035\linewidth]{icon/noun_Box_1650724.png} &

      \includegraphics[width = 0.035\linewidth]{icon/noun_dish_washer_3307528.png} &
      \includegraphics[width = 0.035\linewidth]{icon/noun_Kettle_3002541.png}  &
      \includegraphics[width = 0.035\linewidth]{icon/eyeglasses.png} &
       \includegraphics[width = 0.035\linewidth]{icon/lighter.png} &
      \includegraphics[width = 0.035\linewidth]{icon/noun_Microwave_1041630.png} &
      \includegraphics[width = 0.035\linewidth]{icon/pliers.png} &
      \includegraphics[width = 0.035\linewidth]{icon/noun_Table_59987.png}
      \vspace{-0.2mm}
     \\

\midrule
\ours-Percep + GT-Act$^*$ & 79.5 & 79.6 & 92.5 & 94.6 & 91.2 & 94.7 & 82.3 & 87.3 & 73.4 & 80.4 &  71.5 & 87.9 & 92.2 & 86.4 & 92.9 & 83.8 & 78.7 \\ \midrule
Ditto\cite{jiang2022ditto}+UMP-Act\cite{xu2022umpnet} & 30.9 & 37.0 & 43.8 & 40.3 & 52.1 & 36.6 & 40.8 & 44.7 & 43.7 & 42.5 & 52.2 & 37.3 & 30.7 & 41.7 & 52.1 & 39.3 & 30.0  \\ 
Ditto\cite{jiang2022ditto}+\ours-Act & 24.4 & 40.4 & 48.2 & 43.6 & 36.0 & 66.6 & 43.4 & 50.8 & 70.5 & 52.5 & 54.0 & 41.2 & 33.0 & 42.1 & 60.9 & 36.6 & 31.4 \\  
\ours-NoHistory & 48.8 & \textbf{75.9} & 86.1 & 82.4 & 66.1 & 89.8 & \textbf{68.4} & \textbf{86.7} & 64.6 & 87.5 & \textbf{95.6} & 69.8 & \textbf{49.6} & \textbf{62.7} & {83.9} & \textbf{72.1} & 43.6 \\
\ours  & \textbf{71.5} & 70.1 & \textbf{93.1} & \textbf{87.0} & \textbf{68.9} & \textbf{92.2} & 61.6 & 85.2 & \textbf{75.0 } & \textbf{95.7} &{ 89.1} & \textbf{78.8} & 49.1 & 58.6 & \textbf{85.3} & 67.0 & \textbf{49.3} \\ \bottomrule
\end{tabular}
\vspace{-2mm}
\vspace{-3mm}
}
\end{table*}

\begin{table*}[t]
\footnotesize
\vspace{-2mm}
\caption{\textbf{Joint evaluation.}}
\label{tab:join}
\vspace{-2mm}
\centering
{
    \setlength\tabcolsep{2 pt}
    \begin{tabular}{l|lcccccccc|cccccc|cc|cc|c}
    \toprule
     & \multicolumn{15}{c|}{Revolute joint}  & \multicolumn{4}{c|}{Prismatic joint}  \\ 
     & \multicolumn{9}{c|}{Unseen Instances in Training Categories } &  \multicolumn{6}{c|}{Unseen Categories} & \multicolumn{2}{c|}{Unseen Ins.} &  \multicolumn{2}{c|}{Unseen Cat.} & Type \\ 
     & 
      \includegraphics[width = 0.025\linewidth]{icon/noun_Cabinet_2881254.png} &
      \includegraphics[width = 0.025\linewidth]{icon/noun-scissor-4857934.png} &
      \includegraphics[width = 0.025\linewidth]{icon/noun_safe_1202915.png} &
      \includegraphics[width = 0.025\linewidth]{icon/folding_knife.png} &
      \includegraphics[width = 0.025\linewidth]{icon/noun_Laptop_2291662.png} &
      \includegraphics[width = 0.025\linewidth]{icon/noun_Box_1650724.png} &
      \includegraphics[width = 0.025\linewidth]{icon/noun_Fridge_1875643.png} &
      \includegraphics[width = 0.025\linewidth]{icon/noun_folding_chair_2151184.png} &
      \includegraphics[width = 0.025\linewidth]{icon/noun_trashcan_2244926.png} &
      
      \includegraphics[width = 0.025\linewidth]{icon/noun_Microwave_1041630.png} &
      \includegraphics[width = 0.025\linewidth]{icon/noun_dish_washer_3307528.png} &
      \includegraphics[width = 0.025\linewidth]{icon/lighter.png} &
      \includegraphics[width = 0.025\linewidth]{icon/eyeglasses.png} &
      \includegraphics[width = 0.025\linewidth]{icon/pliers.png} &
      \includegraphics[width = 0.025\linewidth]{icon/noun_Kettle_3002541.png} & 
      
      \includegraphics[width = 0.025\linewidth]{icon/noun_Cabinet_2881254.png} &
      \includegraphics[width = 0.025\linewidth]{icon/noun_kitchen_pot_3363643.png} &
      \includegraphics[width = 0.025\linewidth]{icon/noun_dish_washer_3307528.png} &
      \includegraphics[width = 0.035\linewidth]{icon/noun_Table_59987.png} &
      mAcc 
      \vspace{-0.2mm}
     \\
    \midrule
    & \multicolumn{19}{c|}{ Rotation error (in degree)  $\downarrow$} & $\uparrow$ \\\midrule
    Heuristic  & 40.0   & 89.8  & 75.1 & 89.2  & 15.4  & 10.4  & 47.8 & 59.64 & 9.17 & 81.7  & 40.5 & {88.3}   & {84.8}  & {89.2} & {79.4} & {56.8} & {85.9}  & {81.9} & {69.7} & 52.7 \\
    Ditto \cite{jiang2022ditto} & 0.83  & 0.82  & 12.7  & 3.17 & 15.6 & 32.8 & \textbf{0.36} & 75.83 & 89.63 & 0.76 &  \textbf{3.02} & \textbf{1.20} & 8.08 & 2.98 &  35.6 & 85.4 & 3.63 & 2.93 & \textbf{1.27} & 68.9 \\
    \ours  & \textbf{0.39}  & \textbf{0.79 } & \textbf{11.43} & \textbf{1.02}  & \textbf{5.42 } & \textbf{8.61 } & {0.44} & \textbf{2.11} & \textbf{3.72} & \textbf{0.49} & 3.74  & {1.77}   & \textbf{7.52}  & \textbf{1.94}  & \textbf{35.3 }& \textbf{1.49}  & \textbf{0.27} & \textbf{2.82} & 3.34 & \textbf{86.7} \\
    \midrule
    & \multicolumn{15}{c|}{ Position error  for revolute joint (in normalized scale) $\downarrow$} &  \multicolumn{5}{c}{\cellcolor{grey}} \\\midrule
    Heuristic  & 0.79   & 0.76  & 0.71 & 0.51  & 0.67  & 0.46  & 0.65  & 0.57 & 0.48 & 0.82 & 0.67  & 0.76   & 1.17  & 0.73  & 0.42&  \multicolumn{5}{c}{\cellcolor{grey}}  \\
    Ditto \cite{jiang2022ditto} & 0.22   & 0.61 & 0.19 & 0.37   & \textbf{0.14} & 0.23 & 0.25 & 0.32 & 0.44 & 0.34 & 0.39 & \textbf{0.13} & 0.77 & 0.46 & 1.05 &  \multicolumn{5}{c}{\cellcolor{grey}} \\
    \ours & \textbf{0.06}   & \textbf{0.12}  & \textbf{0.03} & \textbf{0.26}  & {0.24}  & \textbf{0.07}  & \textbf{0.07 }   & \textbf{0.01} & \textbf{0.05} &  \textbf{0.04}  &\textbf{ 0.05}  & {0.17} & \textbf{0.41}  & \textbf{0.13} & \textbf{0.41}  &  \multicolumn{5}{c}{\cellcolor{grey}} \\
    \bottomrule
    \end{tabular}
}
\vspace{-5mm}
\end{table*} 

\section{Experiments}
\label{sec:exp}

We train \emph{single} perception and interaction models and evaluate them on 48 unseen instances from 10 categories and 77 instances from 7 unseen categories chosen from PartNet-Mobility.
When evaluating our method in simulation, an agent executes actions directly in our PyBullet \cite{coumans2016pybullet} environment. For the real-world proof of concept, we generate qualitative results for the perception component of our pipeline.

\mypara{Real-world setup.}
To demonstrate the feasibility of \ours in the real-world settings, we set up a single-arm tabletop environment, as shown in Fig.~\ref{fig:result_real}. The robot arm is equipped with a cylindrical pusher, which moves the object parts based on the inferred actions. The environment has four Intel RealSense RGBD cameras, together capturing a RGB point cloud of the object. The following video shows the real-world pipeline: \url{https://sfa.cs.columbia.edu}.

\mypara{Metrics.}
To better understand the quantitative performance of \ours against competing algorithms, we measure various metrics in simulation.
We first evaluate the the effectiveness of the interaction policies independent of the perception model by measuring the \emph{optimal action ratio}, which is \# optimal action / \# total action~\cite{Gadre2021ActTP}.
An action is optimal if it successfully moves a part that has not been discovered.
If all parts are discovered, moving any part is considered optimal.

The performance of object structure discovery is measured by following two aspects: 1) \emph{Part segmentation and reconstructions.}  Evaluated by part-wise 3D Intersection over Union (IoU) between predicted and ground truth part geometry. 2) \emph{Joint inference.} The accuracy of joint estimation is evaluated by 1) classification accuracy (between prismatic or revolute). 2) axis orientation error in degree. 3) axis position error in normalized scale (revolute joint only). All objects are scaled to fit in a $2\times 2 \times 2$ cube in this dataset, and position error is evaluated with respect to this scale.

\mypara{Baselines and Ablations.}
We test and compare with the following alternative interaction or perception module to study the efficacy of our system design: 
\begin{itemize}[leftmargin=3mm]

\item \emph{GT-Act (Oracle)}: to evaluate the perception module's performance upper bound, we test our perception module with optimal actions computed based on the ground truth state.

\item \emph{UMP-Net} \cite{xu2022umpnet}: an interaction policy that aims to change an objects' joint state.

\item \emph{Ditto} \cite{jiang2022ditto}: a perception network that infers object's part segmentation and joint parameters from a single-step interaction. We combine Ditto with the other interaction policies to form a full pipeline.  

\item \emph{Heuristic}: Heuristic baseline for joint inference with ICP. Details can be found in Supp.   

\item \emph{AtP} \cite{Gadre2021ActTP}: An interaction and perception model, which considers only 2D sequential action and 2D part segmentation. 

\item \emph{NoHistory}: An ablated version of \ours to evaluate the perception module's performance when multi-step part aggregation is not used as input for interaction or perception.

\end{itemize}

\subsection{Experimental Results} %

\begin{figure}[t]
\includegraphics[width=0.47\linewidth]{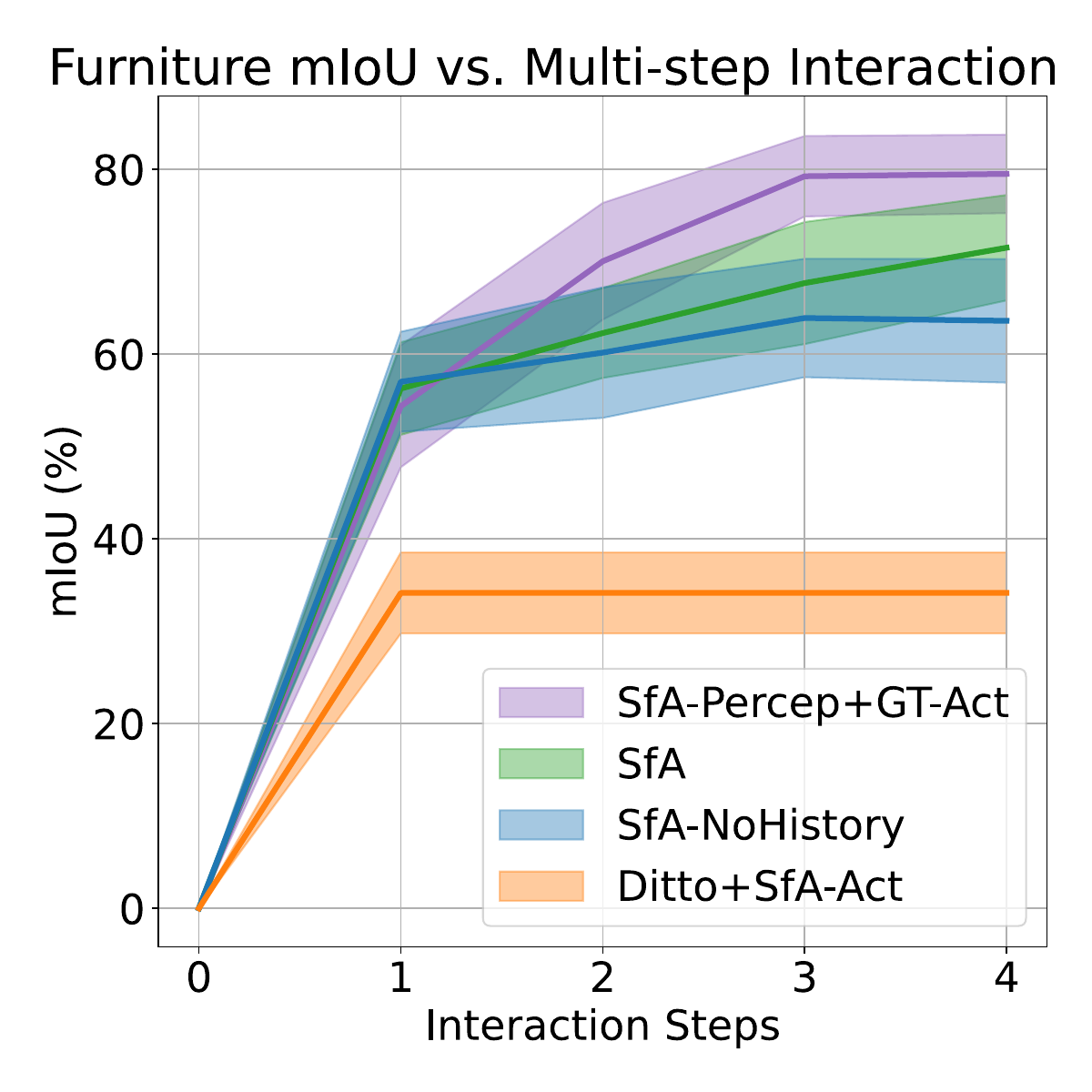}~ 
\includegraphics[width=0.47\linewidth]{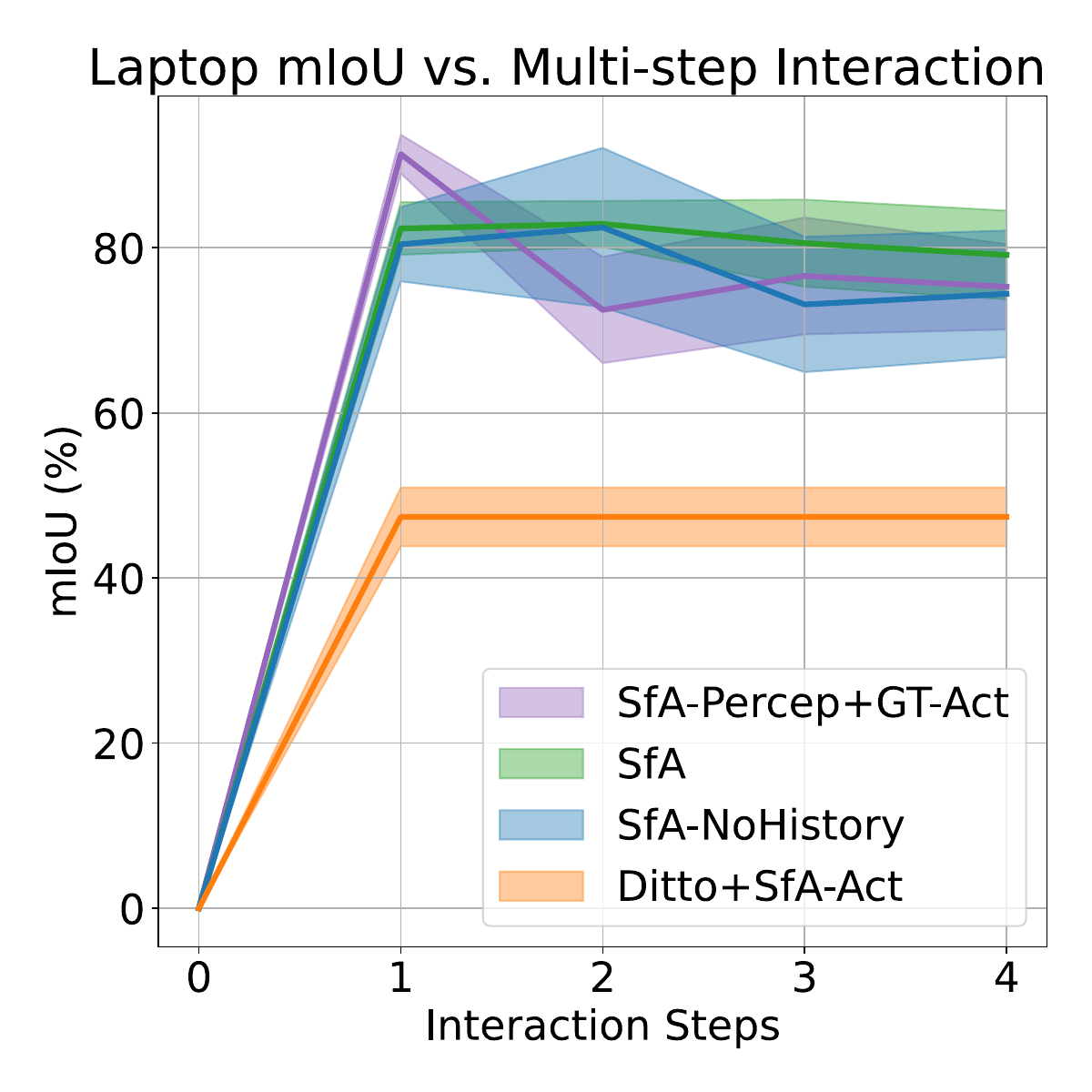} \vspace{-5mm}
\caption {\footnotesize{\textbf{IoU w.r.t steps.} \ours can better discover parts with sequential interactions compared to single-step baseline [Ditto+SfA-Act]\cite{jiang2022ditto}.}, especially on multi-part objects such as furniture. \ours can discover the full structure of two-part objects in one interaction step.}
\label{fig:results-miou-steps} 
\vspace{-5mm}
\end{figure}

\vspace{-2mm}
\mypara{\ours outperforms baseline pipelines made of state-of-the-art models.}
SfA goes beyond combining state-of-the-art components; Tab.~\ref{tab:part-discovery} illustrates this empirically. SfA, on average, outperforms the combination of existing interaction and perception modules (Ditto+UMP-Act) by over 25 percentage points on the 3D reconstruction task with unseen objects. This result also suggests the immense benefit of considering perception and interaction \emph{in conjunction} (i.e., interaction is based on perception and vice versa).

\mypara{Generalization to unseen objects and categories.} Our method makes no category-level assumptions, and allows it to generalize across categories. Tab.~\ref{tab:action},~\ref{tab:part-discovery}, ~\ref{tab:join}, show that \ours is able to achieve similar performance on unseen categories when compared to training categories, and outperforms alternative methods for the majority of the categories. Specifically, the \ours interaction model beats the closest baseline by 8 optimal action points on unseen categories.
For objects with novel kinematics structures such as glasses, the pipeline performance is slightly worse than categories such as microwave, but still outperforms the best competing methods by 16 percentage points in the mIoU evaluation as seen in Tab \ref{tab:part-discovery}. 

\mypara{3D actions are necessary.} 
Observing AtP's performance in Tab.~\ref{tab:action}, we see that while 2D action space is sufficient for simple objects like scissors, it is not effective for complex objects with different joint types, and results in close to zero effective actions for many object categories. The AtP baseline's performance drops considerably when the object cannot fully be observed from the top-down view. In contrast, our interaction policy is able to effectively infer informative 3D actions for a wide variety of objects. Furthermore, we also compare extensively against baselines that employ 3D continuous action spaces.
Specifically, we compare to baselines that employ UMP-Net (see Tab.~\ref{tab:action} and Tab.~\ref{tab:part-discovery}). \ours outperforms the 3D action space baselines in nearly all categories for action inference (Tab.~\ref{tab:action}) and parts segmentation (Tab.~\ref{tab:part-discovery}).

\mypara{Sequential interaction boosts performance.}  
Based on the results in Fig.~\ref{fig:results-miou-steps}, we can observe that our method can not only discover new parts, but also segment parts better than Ditto~\cite{jiang2022ditto}, a single-step interaction baseline as well as our ablated \ours-NoHistory baseline. The improvement is more salient for objects with more than two parts (e.g., furniture and refrigerators). 
Comparing \ours and \ours-Perception (Percep.) + GT-Act in Tab.~\ref{tab:part-discovery}, \ours is competitive with the ablated version with GT interactions.
This result indicates the relative strength of the interaction module the pipeline. 

\mypara{Learned history aggregation helps.}  By using informative interactions and aggregating visual observations in 3D,  \ours could reveal and track surfaces that are initially occluded and better reconstruct part geometry (e.g., the inside of a drawer). 

Comparing \ours and \ours-NoHistory in Tab.~\ref{tab:part-discovery}, we see that for most categories the addition of history improves performance. These gains are most pronounced for objects with more than three parts. In certain two-part object categories, the NoHistory baseline beats \ours. 
This may be caused by the accumulation of perception errors in the multi-step part aggregation process. 

\mypara{\ours generalizes to real-world data.}
To validate the generalization of our approach to real-world data, we implement a robot system that uses a 6DoF robot arm, UR-5, and four RealSense cameras to capture registered RGBD images of real-world articulated objects.
We deploy the full SfA---trained in simulation---directly on this hardware, executing actions sequentially in accordance with the inferred action, recovering structure along the way.
Fig.~\ref{fig:result_real} demonstrates part and joint discovery and part tracking.
These results validate the feasibility of \ours to recover CAD models from real-world RGBD observations. 

\mypara{Limitation and assumptions.}
Our pipeline assumes that only one joint is activated at each interaction step.
While this assumption is mainly satisfied by our learned interaction policy, there can still be cases violating this assumption. 

Additionally our algorithm does not estimate parameters like friction, which can be useful for robot manipulation. 

\section{Conclusion}
\label{sec:conclusion}

We present \ours, a learning framework that discovers 3D parts geometry and joint parameters of novel articulated objects through a sequence of inferred interactions.
Our results show that by coupling interactions and perception, the model can discover and reconstruct 3D articulated CAD models of objects from novel categories and with unknown kinematic structures. 
These results substantiate \ours's potential to enable robots to interact and reconstruct 3D articulated CAD models
autonomously.

{\small \mypara{Acknowledgements.}
This work was supported in part by NSF Awards \#2143601, \#2037101, and \#2132519. We would like to thank Google for the UR5 robot hardware.  
SYG is supported by a NSF Graduate Research Fellowship.
The views and conclusions contained herein are those of the authors and should not be
interpreted as necessarily representing the official policies, either expressed or implied, of the sponsors.
}


{\small
\bibliographystyle{IEEEtranS}
\bibliography{egbib}
}

\end{document}